\documentclass{article}
\usepackage{arxiv}
\usepackage{microtype}
\usepackage{graphicx}
\usepackage{booktabs} 
\usepackage{amsmath}
\usepackage{bbm}

\usepackage{nicefrac}  
\usepackage{lipsum}
\usepackage{graphicx}
\usepackage{amsfonts}
\usepackage{mathabx}
\usepackage[natbibapa]{apacite}
\usepackage{wrapfig}

\usepackage{xcolor}
\usepackage[colorlinks=true,citecolor=red]{hyperref}   

\usepackage[utf8]{inputenc} 
\usepackage[T1]{fontenc}    
\usepackage{hyperref}       
\usepackage{url}            
\usepackage{booktabs}       
\usepackage{amsfonts}       
\usepackage{amsmath}
\usepackage{nicefrac}       
\usepackage{microtype}      
\usepackage{lipsum}
\usepackage{graphicx}

\usepackage{graphicx}
\usepackage[ruled,vlined]{algorithm2e}
\usepackage{subcaption}

\newcommand{\ro}{\mathbf{o}}
\newcommand{\rs}{\mathbf{s}}
\newcommand{\ra}{\mathbf{a}}

\newcommand{\optimal}{\mathcal{O}}

\newcommand{\KL}{D_\mathrm{KL}}

\begin{document}
\title{On the Relationship of Active Inference and Control as Inference}
%
%
\author{
 Beren Millidge \\
  School of Informatics\\
  University of Edinburgh\\
  \texttt{beren@millidge.name}
   \And
 Alexander Tschantz \\
  Sackler Center for Consciousness Science\\
  School of Engineering and Informatics\\
  University   Sussex \\
  \texttt{tschantz.alec@gmail.com} \\
  \And
  Anil K Seth \\
  Sackler Center for Consciousness Science\\
   Evolutionary and Adaptive Systems Research Group\\
  School of Engineering and Informatics\\
  University of Sussex\\
  \texttt{A.K.Seth@sussex.ac.uk} \\
  \And
 Christopher L Buckley \\
  Evolutionary and Adaptive Systems Research Group\\
  School of Engineering and Informatics\\
  University of Sussex\\
  \texttt{C.L.Buckley@sussex.ac.uk} 
  }
%
%
%
\maketitle              
\begin{abstract}
	Active Inference (AIF) is an emerging framework in the brain sciences which suggests that biological agents act to minimise a variational bound on model evidence. Control-as-Inference (CAI) is a framework within reinforcement learning which casts decision making as a variational inference problem. While these frameworks both consider action selection through the lens of variational inference, their relationship remains unclear. Here, we provide a formal comparison between them and demonstrate that the primary difference arises from how value is incorporated into their respective generative models. In the context of this comparison, we highlight several ways in which these frameworks can inform one another.
	
\end{abstract}

Active Inference (AIF) is an emerging framework from theoretical neuroscience which proposes a unified account of perception, learning, and action \citep{friston2006free,friston2008hierarchical,friston2010free}.
This framework posits that agents embody a generative model of their environment and perception and learning take place through a process of variational inference on this generative model by minimizing an information-theoretic quantity -- the variational free energy \citep{friston2017graphical,wainwright2008graphical,beal2003variational}. 
Within this framework, action selection can be cast as a process of inference, underwritten by the same mechanisms which perform perceptual inference and learning \citep{friston2009reinforcement,friston2017active,tschantz2020reinforcement}.
Implementations of active inference have a degree of biological plausibility \citep{friston2017active} and are supported by considerable empirical evidence \citep{walsh2020evaluating}.
Moreover, recent work has shown that active inference can be applied to high-dimensional tasks and environments \citep{millidge2019combining,millidge2020deep,millidge2019implementing,tschantz2019scaling,tschantz2020reinforcement,fountas2020deep,ueltzhoffer2018deep}.

The field of reinforcement learning (RL) \citep{sutton2018reinforcement} 
is also concerned with understanding adaptive action selection. RL assumes that agents seek to maximise the expected sum of rewards (which are generally assumed to be exogenous, not intrinsic to the agent), and then will select the actions that will maximize reward. In recent years, the framework of control as inference (CAI) \citep{levine2018reinforcement,rawlik2010approximate,abdolmaleki2018maximum,attias2003planning,rawlik2013probabilistic} has recast the problem of RL in the language of variational inference. Instead of maximizing rewards, agents must infer actions that lead to optimal trajectories.
This reformulation enables the use of powerful inference algorithms in RL, while also providing a natural method of exploration \citep{haarnoja2018soft,abdolmaleki2018maximum,haarnoja2018applications}. 

Both AIF and CAI view adaptive action selection as a problem of inference.
However, despite these similarities, the formal relationship between the two frameworks remains unclear.
In this work, we attempt to shed light on this relationship.
We present both AIF and CAI in a common language, highlighting connections between the frameworks which may have otherwise been overlooked.
We then consider the key distinction between the frameworks, namely, how `value' or `goals' are encoded into the generative model. 
We discuss how this distinction leads to subtle differences in the objectives that both schemes optimize, and suggest how these differences may impact behaviour.

\section{Formalism}

\begin{figure}[ht]
\vspace{-0.3cm}
\label{pc_rnn_lstm_results_figure}
\hspace{-0.6cm}
\begin{subfigure}{.5\textwidth}
  \centering
  \includegraphics[width=1\linewidth]{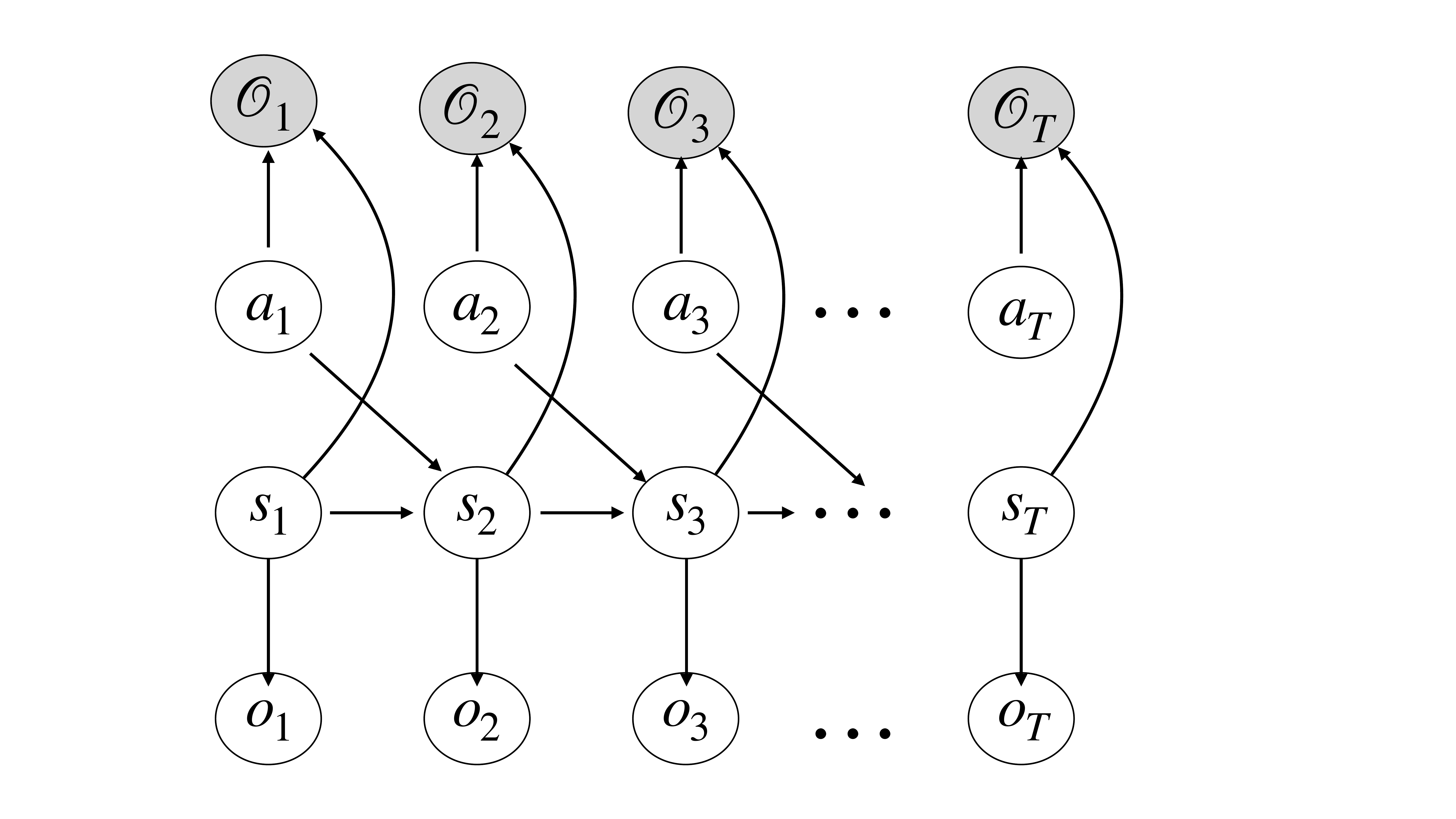}  
  \caption{Control-as-Inference}
\end{subfigure}
\begin{subfigure}{.5\textwidth}
  \centering
  \includegraphics[width=1\linewidth]{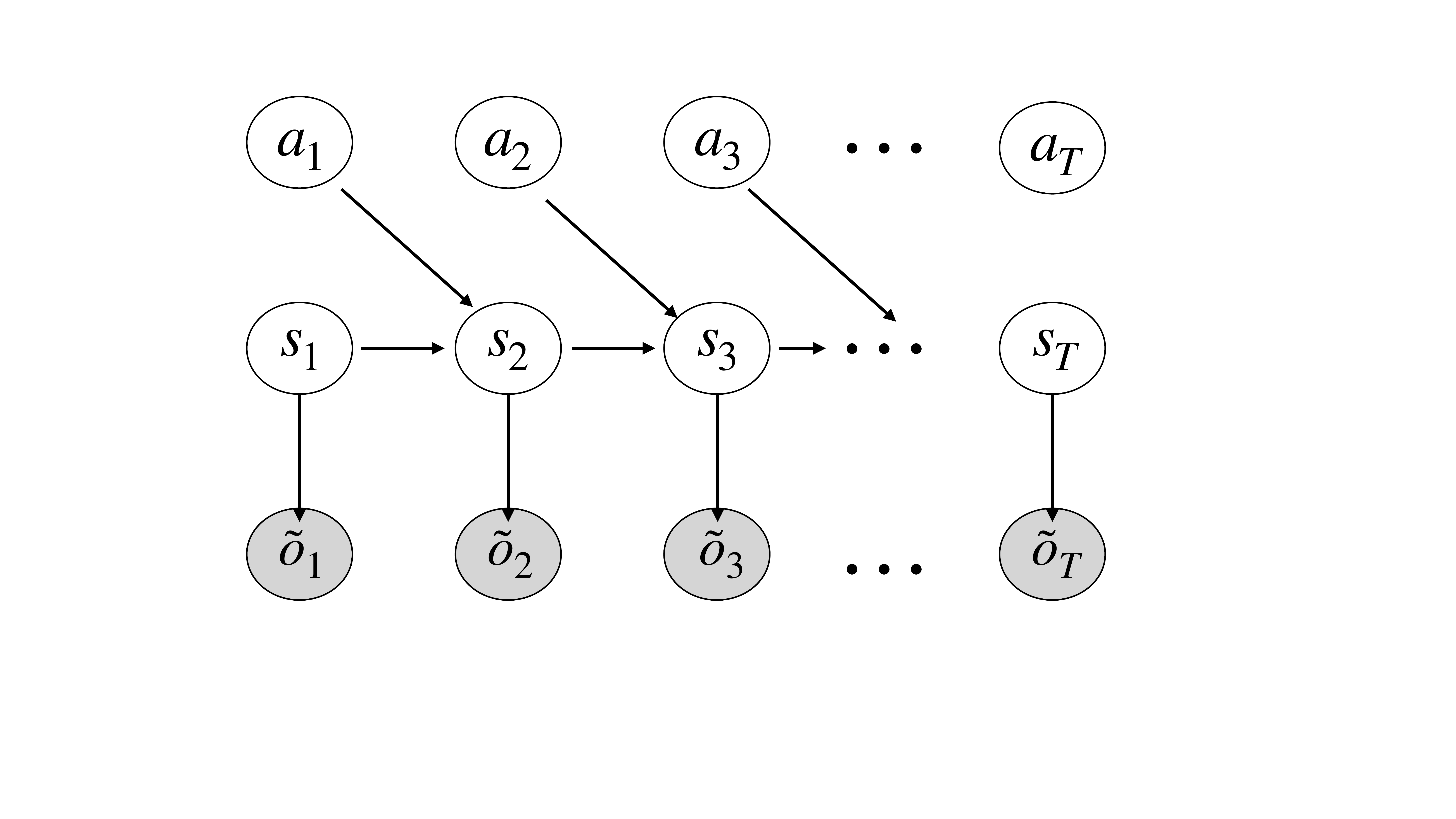}  
  \caption{Active Inference}
\end{subfigure}
\end{figure}
Both AIF and CAI can be formalised in the context of a partially observed Markov Decision Process (POMDP).
Let $\ra$ denote actions, $\rs$ denote states and $\ro$ denote observations. 
In a POMDP setting, state transitions are governed by $\rs_{t+1} \sim p_{\texttt{env}}(\rs_{t+1}|\rs_{t}, \ra_{t})$ whereas observations are governed by $\ro_t \sim p_{\texttt{env}}(\ro_t | \rs_t)$.
We also assume that the environment possesses a reward function $r: \mathcal{S} \times \mathcal{A} \rightarrow \mathbb{R}^1$ which maps from state-action pairs to a scalar reward. 
Agents encode (and potentially learn) a generative model $p(\rs_{t:T}, \ra_{t:T}, \ro_{t:T})$ that describes the relationship between states, actions, observations up to a time horizon $T$.
AIF and CAI are both concerned with inferring the posterior distribution over latent variables $p(\ra_{t:T}, \rs_{t:T}|\ro_{t:T})$.
However, solving this `value-free' inference problem will not by itself lead to adaptive behaviour.
Additional assumptions are required to bias inference towards inferring actions that lead to `valuable' states.

\section{Control as Inference}

CAI incorporates the notion of `value' by introducing an additional `optimality' variable $\mathcal{O}_t$, where $\mathcal{O}_t = \mathbf{1}$ implies that time step $t$ was optimal, meaning that given that the later timesteps $t+1:T$ are also optimal, the whole trajectory $t:T$ is optimal. 
In what follows, we simplify notation by assuming $p(\mathcal{O}_t) = p(\mathcal{O}_t = \mathbf{1})$.
The goal of CAI then to recover the posterior over states and actions, given the belief that the agent will observe itself being optimal, i.e. $p(\rs_t, \ra_t|\ro_t, \mathcal{O}_t)$. By including the optimality variable we can write the agent's generative model as $p(\rs_{t:T}, \ra_{t:T}, \ro_{t:T}, \optimal_t) = \prod_t^T p(\optimal_t | \rs_t, \ra_t)p(\ro_t | \rs_t)p(\ra_t | \rs_t)p(\rs_t | \rs_{t-1},\ra_{t-1})$ \footnote{Note that CAI is usually formulated in the context of an MDP rather than a POMDP. We have presented the POMDP case to maintain consistency with AIF, but both frameworks can be applied in both MDPs and POMDPs.}.
Inferring the posterior $p(\rs_t, \ra_t|\ro_t, \mathcal{O}_t)$ is generally intractable, but it can approximated by introducing an auxillary variational distribution $q_\phi(\rs_t, \ra_t) = q_\phi(\ra_t|\rs_t)q(\rs_t)$ and optimising the variational bound $\mathcal{L}(\phi)$:
\begin{equation}
\label{eq:cai-bound}
\begin{aligned}
    \mathcal{L}(\phi) &= \KL \Big(q_\phi(\rs_t, \ra_t) \Vert p(\rs_t, \ra_t,\ro_t,\optimal_t) \Big) \\
    &= \underbrace{-\mathbb{E}_{q_\phi(\rs_t, \ra_t)}\big[ \ln p(\optimal_t | \rs_t, \ra_t) \big]}_{\texttt{Extrinsic Value}} + \underbrace{\KL \Big( q(\rs_t) \Vert p(\rs_t | \rs_{t-1},\ra_{t-1}) \Big)}_{\texttt{State divergence}} \\
    & \ \ + \underbrace{\mathbb{E}_{q(\rs_t)} \big[\KL \Big( q_\phi(\ra_t | \rs_t) \Vert p(\ra_t | \rs_t) \Big) \big]}_{\texttt{Action Divergence}} - \underbrace{\mathbb{E}_{q_\phi(\rs_t, \ra_t)}\big[ \ln p(\ro_t | \rs_t)\big]}_{\texttt{Observation Ambiguity}} \\
\end{aligned}
\end{equation}
Minimising Eq. \ref{eq:cai-bound} -- a process known as variational inference -- will cause the approximate posterior $q_\phi(\rs_t, \ra_t)$ to tend towards the true posterior $p(\rs_t, \ra_t|\ro_t, \mathcal{O}_t)$, and will also cause the marginal-likelihood of optimality $p(\mathcal{O}_t)$ to be maximised. 

The second equality in Eq. \ref{eq:cai-bound} demonstrates that this variational bound can be decomposed into four terms.
The first term (extrinsic value) quantifies the likelihood that some state-action pair is optimal. 
In the CAI literature, the likelihood of optimality is usually defined as $p(\optimal_t | \rs_t, \ra_t) := e^{r(\rs_t, \ra_t)}$, such that $\ln p(\optimal_t | \rs_t, \ra_t) = r(\rs_t, \ra_t)$.   
Extrinsic value thus quantifies the expected reward of some state-action pair, such that minimising $\mathcal{L}(\phi)$ maximises expected reward. \footnote{An additional, but minor difference between the frameworks is that CAI typically assumes that rewards come from state-action pairs $r \sim r(\rs_t, \ra_t)$ while AIF typically assumes rewards are a function of observations $r \sim r(\ro_t)$. This difference can be straightforwardly finessed by either reparametrising CAI-rewards in terms of observations or AIF-rewards in terms of states and actions.} The state divergence and action divergence terms quantify the degree to which states and actions diverge from their respective priors. 
The approximate posterior over states and the agent's model of state dynamics are assumed to be equal $q(\rs_t) := p(\rs_t | \rs_{t-1}, \ra_{t-1})$, such that the agent believes it has no control over the dynamics except through action. If this assumption is not made, this typically leads to risk-seeking policies, as the agent assumes it can alter the dynamics arbitrarily to avoid bad outcomes \citep{levine2018reinforcement}. This assumption eliminates the second term (state divergence) from the bound.
Moreover, under the assumption that the action prior is uniform $p(\ra_t | \rs_t) := \frac{1}{|\mathcal{A}|}$, the action divergence term reduces to the negative entropy of actions. 
Maximising an action entropy term provides several benefits, including a mechanism for offline learning, improved exploration and increased algorithmic stability \citep{haarnoja2018soft,haarnoja2018applications}. 
The fourth term (observation ambiguity) encourages agents to seek out states which have a precise mapping to observations, and only arises in a POMDP setting. In effect this leads to agents that implicitly try to minimize the overhead of a POMDP compared to an MDP by trying to stay within regions of the state-space which have a low-entropy mapping to observations.

\subsection{Inferring plans}
Traditionally, CAI has been concerned with inferring \emph{policies}, or time-dependent state-action mappings. 
Here, we reformulate the standard CAI approach to instead infer fixed action sequences, or \emph{plans} $\pi = \{\ra_t, ... , \ra_T\}$. 
Specifically, we derive a novel variational bound for CAI and show that it can be used to derive an expression for the optimal time-independent plan.
We adapt the generative model and approximate posterior to account for a temporal \emph{sequence} of variables $p(\rs_{t:T}, \pi, \ro_{t:T}, \optimal_{t:T}) = \prod_{t}^T p(\optimal_t | \rs_t, \pi)p(\ro_t | \rs_t)p(\rs_t | \rs_{t-1}, \pi)p(\pi)$ and $q(\rs_{t:T}, \pi) = \prod_{t}^T q(\rs_t|\pi)q(\pi)$.
The optimal policy can then be retrieved as:

\begin{equation}
\begin{aligned}
    \mathcal{L} &= \KL \Big( q(\rs_{t:T}, \pi) \Vert p(\rs_{t:T}, \pi, \ro_{t:T}, \optimal_{t:T}) \Big) \\
    &= \KL \Big( q(\pi)\Vert p(\pi) \exp(- \sum_t^T \mathcal{L}_t(\pi) ) \Big) 
    \implies q^*(\pi) = \sigma \Big(p(\pi) - \sum_t^T \mathcal{L}_t(\pi)\Big) 
\end{aligned}
\end{equation}
The optimal policy is thus a path integral of the $\mathcal{L}_t(\pi)$ which can be written as:
\begin{equation}
\begin{aligned}
    \mathcal{L}_t(\pi) &= \mathbb{E}_{q(\rs_t | \pi)}\big[ \ln q(\rs_t|\pi) - \ln p(\rs_t, \pi, \ro_t, \optimal_t)] \\
    &= -\underbrace{\mathbb{E}_{q(\rs_t | \pi)}\big[ \ln p(\optimal_t | \rs_t, \pi) \big]}_{\texttt{Extrinsic Value}} + \underbrace{\KL \Big(q(\rs_t | \pi) \Vert p(\rs_t | \rs_{t-1}, \pi) \Big)}_{\texttt{State divergence}} 
     - \underbrace{\mathbb{E}_{q(\rs_t | \pi)}\big[ \ln p(\ro_t | \rs_t) \big]}_{\texttt{Observation Ambiguity}} 
\end{aligned}
\end{equation}
Which is equivalent to Eq. \ref{eq:cai-bound} except that it omits the action-divergence term.

\section{Active Inference}

Unlike CAI, AIF does not introduce additional variables incorporate `value' into the generative model. 
Instead, AIF assumes that the generative model is intrinsically biased towards valuable states or observations. 
For instance, we might assume that the prior distribution over observations is biased towards observing rewards, $\ln \tilde{p}(\ro_{t:T}) \propto e^{r(\ro_{t:T})}$, where we use notation $\tilde{p}(\cdot)$ to denote a biased distribution\footnote{AIF is usually formulated only in terms of observations where some observations are more desired than others. We introduced rewards to retain consistency with CAI.}.
Let the agent's generative model be defined as $\tilde{p}(\rs_{t:T}, \ro_{t:T}, \pi) = p(\pi)\prod_{t}^{T} p(\rs_t | \ro_t, \pi)\tilde{p}(\ro_t | \pi)$, and the approximate posterior as $q(\rs_{t:T}, \pi) = q(\pi)\prod_{t}^Tq(\rs_t|\pi)$.

It is then possible to derive an analytical expression for the optimal plan:
\begin{equation}
    \begin{aligned}
    -\mathcal{F}(\pi) &= \mathbb{E}_{q(\ro_{t:T},\rs_{t:T},\pi)}\big[ \ln q(\rs_{t:T},\pi) - \ln \tilde{p}(\ro_{t:T}, \rs_{t:T},\pi) \big]  \\ 
    &\implies q^*(\pi) = \sigma \big( \ln p(\pi) -   \sum_t^T \mathcal{F}_t(\pi)\big) 
\end{aligned}
\end{equation}
where $-\mathcal{F}_t(\pi)$ is referred to as the \emph{expected free energy} (note that other functionals are consistent with AIF \citep{millidge2020whence}). 
Given a uniform prior over policies, behaviour is determined by the expected free energy functional, which decomposes into: 
\begin{equation}
\begin{aligned}
\label{eq:efe}
    -\mathcal{F}_t(\pi) &= -\mathbb{E}_{q(\ro_t, \rs_t|\pi)}\big[ \ln q(\rs_t|\pi) - \ln \tilde{p}(\ro_t, \rs_t|\pi) \big] \\
    &= \underbrace{-\mathbb{E}_{q(\ro_t, \rs_t|\pi)}\big[\ln \tilde{p}(\ro_t|\pi)\big]}_{\texttt{Extrinsic Value}} - \underbrace{\mathbb{E}_{q(\ro_t|\pi)}\Big[ \KL \big( q(\rs_t | \ro_t, \pi) \Vert q(\rs_t|\pi) \big) \Big]}_{\texttt{Intrinsic Value}} 
\end{aligned}
\end{equation}
where we have made the assumption that the inference procedure is approximately correct, such that $q(\rs_t|\ro_t,\pi) \approx p(\rs_t|\ro_t, \pi)$.  
As agents are required to minimise Eq. \ref{eq:efe}, they are required to maximise both extrinsic and intrinsic value. 
Extrinsic value measures the degree to which expected observations are consistent with prior beliefs about favourable observations. 
Under the assumption that $\ln \tilde{p}(\ro_{t:T}) \propto e^{r(\ro_{t:T})}$, this is equivalent to seeking out rewarding observations.
Intrinsic value is equivalent to the expected information gain over states, which compels agents to seek informative observations that most reduce posterior-state uncertainty and which leads to large updates between the prior and posterior beliefs about states. 

\subsection{Inferring policies}

While AIF is usually formulated in terms of fixed action sequences, it can also be formulated in terms of policies (i.e. state-action mappings). 
Let the agent's generative model be defined as $\tilde{p}(\rs_t, \ro_t, \ra_t) = p(\rs_t | \ro_t, \ra_t)p(\ra_t|\rs_t)\tilde{p}(\ro_t | \ra_t)$, and the approximate posterior as $q_{\phi}(\rs_t, \ra_t) = q_{\phi}(\ra_t|\rs_t)q(\rs_t)$.
We can now write the expected free energy functional in terms of the policy parameters $\phi$:
\begin{equation}
\label{eq:policies}
    \begin{aligned}
    -\mathcal{F}_t(\phi) &= \mathbb{E}_{q(\ro_t,\rs_t,\ra_t)}\Big[\ln q_\phi(\ra_t, \rs_t) - \ln \tilde{p}(\rs_t, \ro_t, \ra_t) \Big] \\
    &= -\underbrace{\mathbb{E}_{q(\ro_t|\ra_t)}\big[\ln \tilde{p}(\ro_t|\ra_t) \big]}_{\texttt{Extrinsic Value}} - \underbrace{\mathbb{E}_{q(\ro_t,\ra_t  | \rs_t)}\Big[ \KL \big( q(\rs_t | \ro_t, \ra_t) \Vert q(\rs_t|\ra_t) \big) \Big]}_{\texttt{Intrinsic Value}}  + \underbrace{\mathbb{E}_{q(\rs_t)}\Big[ \KL \big( q_\phi(\ra_t | \rs_t) \Vert p(\ra_t | \rs_t) \big) \Big]}_{\texttt{Action Divergence}} 
\end{aligned}
\end{equation}
Inferring policies with AIF thus requires minimizing an action divergence term which is not present when inferring plans but is directly equivalent to the action-divergence term in the CAI formulation.

\section{Encoding Value}

We have shown that both AIF and CAI can be formulated as variational inference, for both fixed action sequences (i.e. plans) and policies (i.e. state-action mappings). 
We now move on to consider the key difference between these frameworks -- how they encode `value'. AIF encodes value directly into the generative model as a prior over observations, whereas in CAI the extrinsic value is effectively encoded into the likelihood which, by Bayes rule, relates to the prior as $p(\ro_t | \rs_t) = p(\ro_t)\frac{p(\rs_t)}{p(\rs_t | \ro_t)}$. When applied within a KL divergence, this fraction becomes a negative information gain. We elucidate this distinction by modelling a further variant of active inference, which here we call \emph{likelihood-AIF}, where instead of a biased prior over rewards the agent has a biased likelihood $\tilde{p}(\ro_t, \rs_t) = \tilde{p}(\ro_t | \rs_t)p(\rs_t)$. The likelihood-AIF objective functional  $\hat{\mathcal{F}}(\phi)$ becomes:

\begin{align*}
    -\hat{\mathcal{F}_t}(\phi) &= \mathbb{E}_{q_\phi(\rs_t, \ro_t, \ra_t)}\big[\ln q_\phi(\rs_t, \ra_t) - \ln \tilde{p}(\ro_t, \rs_t, \ra_t) \big]  \\
    &=  \underbrace{-\mathbb{E}_{q_\phi(\rs_t, \ra_t)}\big[ \ln \tilde{p}(\ro_t | \rs_t) \big]}_{\texttt{Extrinsic Value}} + \underbrace{\KL \Big( q(\rs_t) \Vert p(\rs_t | \rs_{t-1},\ra_{t-1}) \Big)}_{\texttt{State divergence}}  + \underbrace{\KL \Big( q_\phi(\ra_t | \rs_t) \Vert p(\ra_t | \rs_t) \Big)}_{\texttt{Action Divergence}}
\end{align*}
If we set $\ln \tilde{p}(\ro_t | \rs_t) = \ln p(\optimal_t | \rs_t, \ra_t)$, this is exactly equivalent to the CAI objective in the case of MDPs. The fact that likelihood-AIF on POMDPs is equivalent to CAI on MDPs is due to the fact that the observation modality in AIF is `hijacked' by the encoding of value, and thus effectively contains one less degree-of-freedom compared to CAI, which maintains a separate veridical representation of observation likelihoods. A further connection is that AIF on MDPs is equivalent to KL control \citep{rawlik2010approximate,rawlik2013probabilistic,rawlik2013stochastic,van2010risk}, and the recently proposed state-marginal-matching \citep{lee2019efficient} objectives. We leave further exploration of these similarities to future work. 

\section{Discussion}

In this work, we have highlighted the large degree of overlap between the frameworks of active inference (AIF) and control as inference (CAI) and have explored how each framework encodes value into the generative model,  thereby turning a value-free inference problem into one that can serve the purposes of adaptive action. CAI augments the `natural' probabilistic graphical model with exogenous optimality variables. \footnote{Utilising optimality variables is not strictly necessary for CAI. In the case of undirected graphical models, an additional undirected factor can be appended to each node \citep{ziebart2010modeling}. Interestingly, this approach bears similarities to the procedure adopted in \cite{parr2019generalised}, suggesting a further connection between generalised free energy and CAI.}. 
In contrast, AIF leaves the structure of the graphical model unaltered and instead encodes value into the generative model directly. These two approaches lead to significant differences between their respective functionals. AIF, by contaminating the veridical generative model with value-imbuing biases, loses a degree of freedom compared to CAI which maintains a strict separation between the veridical generative model of the environment and its goals. In POMDPs, this approach results in CAI being sensitive to an `observation-ambiguity' term which is absent in the AIF formulation. Secondly, the different methods for encoding the probability of goals -- likelihoods in CAI and priors in AIF -- lead to different exploratory terms in the objective functionals. Specifically, AIF is endowed with an expected information gain that CAI lacks. AIF approaches thus lend themselves naturally to goal-directed exploration whereas CAI mandates only random, entropy-maximizing exploration.

These different ways of encoding goals into probabilistic models also lend themselves to more philosophical interpretations. CAI, by viewing goals as an additional exogenous factor in an otherwise unbiased inference process, maintains a clean separation between veridical perception and control, thus maintaining the modularity thesis of separate perception and action modules \citep{baltieri2018modularity}. This makes CAI approaches consonant with mainstream views in machine learning that see the goal of perception as recovering veridical representations of the world, and control as using this world-model to plan actions. In contrast, AIF elides these clean boundaries between unbiased perception and action by instead positing that \emph{biased} perception \citep{tschantz2020learning} is crucial to adaptive action. Rather than maintaining an unbiased world model that predicts likely consequences, AIF instead maintains a biased generative model which preferentially predicts our preferences being fulfilled.  Active-inference thus aligns closely with enactive and embodied approaches \citep{baltieri2019generative,clark2015radical} to cognition, which view the action-perception loop as a continual flow rather than a sequence of distinct stages.

We have thus seen how two means of encoding preferences into inference problems leads to two distinct families of algorithms, each optimising subtly different functionals, resulting in differing behaviour. This raises the natural questions of which method should be preferred, and whether these are the only two possible methods.One can imagine explicitly modelling the expected reward, and biasing inferences with priors over the expected reward. Alternatively, agents could maintain desired distributions over all of states, observations, and actions, which would maximize the flexibility in specifying goals intrinsic to the framework. Future research will explore these potential extensions to the framework, their relation to one another, and the objective functionals they induce.

\bibliography{refs}
\end{document}